\documentclass[10pt,twocolumn,letterpaper]{article}

\usepackage{cvpr}              % To produce the CAMERA-READY version
\usepackage{graphicx}
\usepackage{amsmath}
\usepackage{amssymb}
\usepackage{booktabs}
\usepackage{multirow}
\usepackage[T1]{fontenc}
\usepackage{threeparttable}
\usepackage{url}
\usepackage[pagebackref,breaklinks,colorlinks]
{hyperref}
\usepackage[accsupp]{axessibility}  % Improves PDF readability for those with disabilities.

\usepackage[capitalize]{cleveref}
\crefname{section}{Sec.}{Secs.}
\Crefname{section}{Section}{Sections}
\Crefname{table}{Table}{Tables}
\crefname{table}{Tab.}{Tabs.}

\begin{document}

\title{AutoSplice: A Text-prompt Manipulated Image Dataset for Media Forensics}

\author{Shan Jia \hspace{0.7cm} Mingzhen Huang \hspace{0.7cm} Zhou Zhou\hspace{0.7cm} Yan Ju\hspace{0.7cm} Jialing Cai\hspace{0.7cm}  Siwei Lyu\\
University at Buffalo, State University of New York, NY, USA\\
{\tt\small \{shanjia, mhuang3, zzhou38, yanju, jialingc, siweilyu\}@buffalo.edu}
}
\maketitle

%%%%%%%%% ABSTRACT
\begin{abstract}
Recent advancements in language-image models have led to the development of highly realistic images that can be generated from textual descriptions. However, the increased visual quality of these generated images poses a potential threat to the field of media forensics. This paper aims to investigate the level of challenge that language-image generation models pose to media forensics. To achieve this, we propose a new approach that leverages the DALL-E2 language-image model to automatically generate and splice masked regions guided by a text prompt. To ensure the creation of realistic manipulations, we have designed an annotation platform with human checking to verify reasonable text prompts. This approach has resulted in the creation of a new image dataset called {\em AutoSplice}, containing 5,894 manipulated and authentic images. Specifically, we have generated a total of $3,621$ images by locally or globally manipulating real-world image-caption pairs, which we believe will provide a valuable resource for developing generalized detection methods in this area \footnote{The AutoSplice dataset is available from \url{https://github.com/shanface33/AutoSplice_Dataset}}. %which we believe will provide a valuable resource for future research in this area.
%Recent large-scale language-image models are capable of generating impressive images from textual descriptions. The level of realism and authenticity in these images is so high that they can be indistinguishable from authentic ones, and therefore pose a potential threat to the field of media forensics. In this paper, our goal is to investigate how challenging the language-image generation model will be to media forensics. Unlike current image manipulation datasets that rely on random or manual manipulation, we propose to leverage the DALL-E2 language-image model to generate and splice masked regions automatically, guided by a text prompt. An annotation platform with human checking is designed to ensure reasonable text prompts and the creation of realistic manipulations. We then provide a text-prompt manipulated image dataset, named AutoSplice, which contains 5,894 manipulated and authentic images. More specifically, we have synthesized a total of 3,621 images by either locally or globally manipulating real-world image-caption pairs. %\footnote{Datasets and codes will be available at \href{https://github.com/***}{\url{https://github.com/***}.}}. 
The dataset is evaluated under two media forensic tasks: forgery detection and localization. %with lossless and mildly lossy compressed images. 
Our extensive experiments show that most media forensic models struggle to detect the AutoSplice dataset as an unseen manipulation. However, when fine-tuned models are used, they exhibit improved performance in both tasks. %The dataset is available at \url{https://github.com/shanface33/AutoSplice_Dataset}. 
\end{abstract}

%%%%%%%%% BODY TEXT
\section{Introduction}
\label{sec:intro}
The proliferation of digital media and AI technology have made it easier to manipulate and fabricate digital content. In recent years, the rapid development of powerful deep generative models, such as Variational Autoencoders (VAEs)~\cite{ daniel2021soft, kingma2019introduction}, Generative Adversarial Networks (GAN)~\cite{karras2019style, karras2020analyzing}, diffusion-based models~\cite{dhariwal2021diffusion, saharia2022palette}, and the latest large-scale language-image (LLI) models~\cite{gafni2022make, DALLEramesh2021zero, DALLE2ramesh2022hiera, zhou2021lafite, rombach2022high}, have brought new challenges to the authentication of digital media. The generated images have become increasingly realistic and convincing so that it can be difficult for human eyes to discern as artificial~\cite{sinha2021d2c, wang2022gan}. 

\begin{figure}[t]
    \centering
    \includegraphics[width=0.48\textwidth]{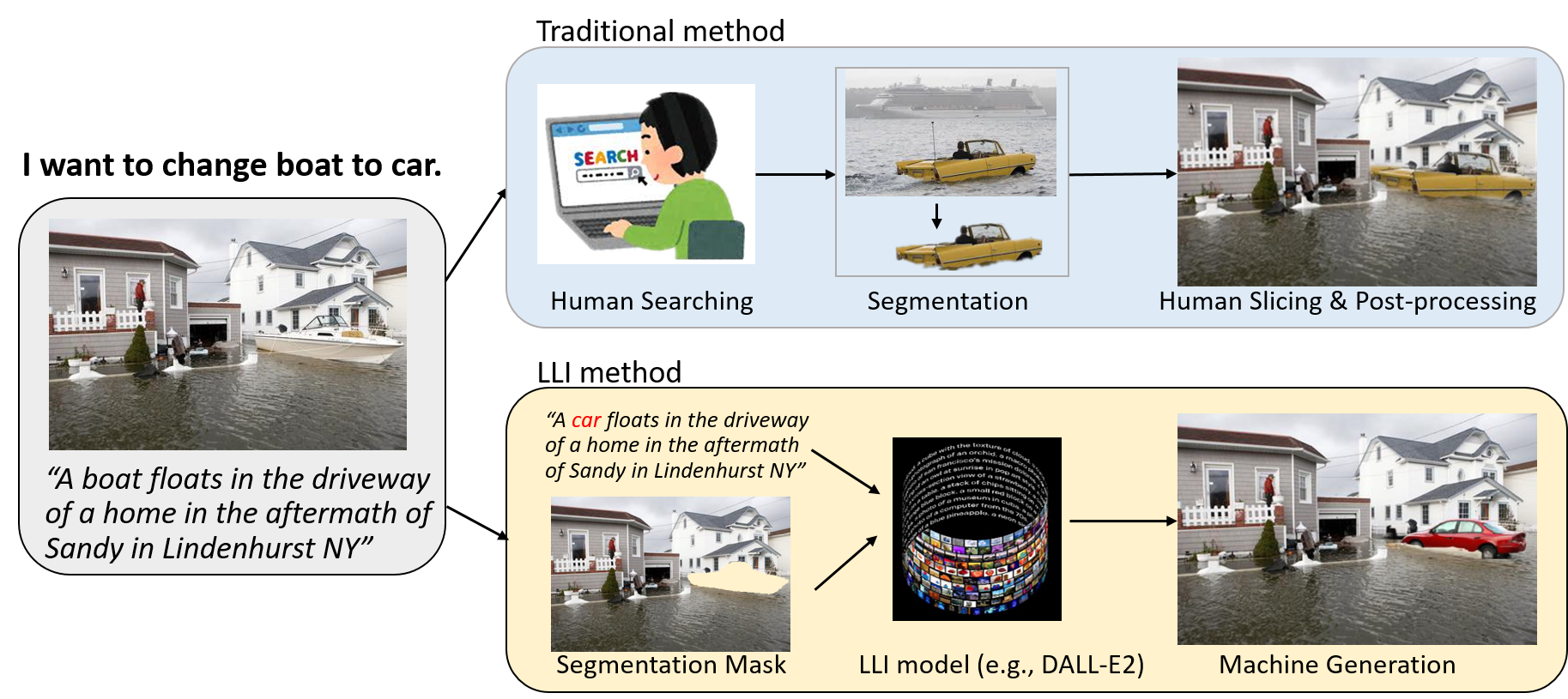}
    \caption{\small Comparison of our text-prompt-based image manipulation pipeline and traditional manual pipeline.}
    \vspace{-0.55cm}
    \label{fig:f1}
\end{figure}

Many efforts~\cite{verdoliva2020media, wang2020cnn, wang2022gan,ju2022fusing} have investigated the challenges of GAN-generated images to media forensics due to the surge of GAN generation models, such as ProGAN~\cite{karras2017progressive}, BigGAN~\cite{brock2018large}, StyleGAN~\cite{karras2019style}, etc. In addition, synthetic images from diffusion models can be identified with high accuracy using similar diffusion models in training, as demonstrated by \cite{corvi2022detection}. The majority of previous studies have focused on entire synthesis using generative models. However, with the emergence of large-scale language-image models, local image manipulations guided by text prompts have become more accessible. Local manipulation of image regions, regardless of size, tends to be more realistic and challenging to detect than entire synthesis. In contrast to local manipulation techniques such as manual copy-move, slicing, or inpainting as described in~\cite{wang2022objectformer}, the latest language and vision models such as DALL-E~\cite{DALLEramesh2021zero, DALLE2ramesh2022hiera} allow for fully automatic and realistic local edits to images guided by a text prompt while preserving semantic information and stylistic elements, as shown in Figure \ref{fig:f1}. These models offer improved generation efficiency, image quality, and content flexibility compared to traditional local manipulation methods, and potentially revolutionize image manipulation.

%Most of these studies focused on entire synthesis based on generative models. The emergence of large-scale language-image models has made it simple to perform local manipulations guided by text prompts. Compared with entire synthesis, local image manipulation trends to be more realistic with partial synthesis in any size of image region and at the mean time, more difficult to detect. 
%Different from local manipulation based on manual copy-move, slicing, or inpainting~\cite{wang2022objectformer} methods, the latest LLI models, like DALL-E~\cite{rombach2022high}, make fully automatic and realistic local edits possible to images, guided by a text prompt while preserve semantic information and stylistic elements. They improve the generation efficiency, image quality, and content flexibility over traditional local manipulation methods, and open up new and exciting opportunities to this field. 

\vspace{-0.2cm}
A key question we investigate in this paper is: how much threat does the state-of-the-art language-image model pose to the current media forensic techniques? To explore this question, we first create a novel text-prompt manipulated image dataset using DALL-E2 \footnote{ \href{https://github.com/lucidrains/DALLE2-pytorch}{\url{https://github.com/lucidrains/DALLE2-pytorch}.}}. 
To create high-quality manipulations, we designed a semi-automatic annotation platform with manual checking. To ensure a diverse set of real-world media data, we utilized caption-image pairs from the Visual News dataset~\cite{liu2020visual} as source data. Using automatic media analysis tools and human annotations, we extracted potential object regions to be manipulated from the images and parsed and replaced the corresponding text prompts in the captions. These captions were then taken as input to the DALL-E2 model for local image manipulation. For each caption-image pair input, we generated manipulated images (three by DALL-E2) and the corresponding manipulation mask. %, as well as two caption-image inconsistency pairs (i.e., the original caption with the generated image and swapped caption with the original image). 
After data cleaning, our AutoSplice dataset contains $3,621$ manipulated images and $2,273$ authentic images.

Our dataset has several advantages over existing relevant media forensics datasets or methods, including high flexibility in content generation, high diversity in manipulation region, and good and reasonable generation quality. We use the large-scale language-image model, DALL-E2, for automatic local manipulation to create realistic forgery images, unlike previous local image manipulation methods based on manual and random object copy-move, slicing, or inpainting. Additionally, we only partially manipulate the image guided by the input region mask instead of generating the entire image, unlike recent semantic editing tools that use diffusion-based LLI models such as DiffEdit~\cite{couairon2022diffedit}, Prompt-to-prompt~\cite{hertz2022prompt}, and Imagic~\cite{kawar2022imagic}. Thanks to the powerful DALL-E2 generation model and human annotations, our dataset contains highly diverse and realistic synthesized images.
We evaluate two media forensics tasks, namely, image forgery detection and image manipulation localization, on the AutoSplice dataset with lossless and lossy compression. Results show that pre-trained methods have limited generalization ability and unreliable prediction in detecting AutoSplice images. Models with fine-tuning on the dataset achieve improved performance, but also show vulnerability to compression.

%-------------------------------------------------------------------------
%-------------------------------------------------------
\section{Related Work}
\label{sec:rel}
\subsection{LLI Synthesis Models}
Recent advancements in attention-based transformer and diffusion models have significantly improved text-to-image generation in the past two years. Several large-scale language-image models have been developed. The DALL-E model~\cite{DALLEramesh2021zero}, proposed by OpenAI in 2021, uses an autoregressive transformer to achieve high-quality image generation on the MS-COCO dataset~\cite{lin2014microsoft} without using any training labels. Other models, such as CogView~\cite{ding2021cogview}, Parti~\cite{yu2022scaling}, and Make-A-Scene~\cite{gafni2022make}, have also trained autoregressive transformer models on text and image tokens for text-to-image generation. In 2022, an updated version of DALL-E, DALL-E2~\cite{DALLE2ramesh2022hiera}, was developed using a diffusion model with CLIP image embeddings, making it computationally more efficient and able to produce higher-quality and more diverse samples. Other models, such as GLIDE~\cite{nichol2021glide}, Stable-Diffusion~\cite{rombach2022high}, and Imagen~\cite{Imagen_saharia2022phot}, have also used diffusion models to improve text-to-image synthesis. Inspired by these powerful LLI synthesis models, several studies have developed text-guided image editing models, including DiffEdit~\cite{couairon2022diffedit}, Prompt-to-prompt~\cite{hertz2022prompt}, Null-text Inversion~\cite{mokady2022null}, Imagic~\cite{kawar2022imagic}, and Muse~\cite{chang2023muse}. These models apply local semantic editing to an image given a text input (with the desired edit) and an optional scene layout (segmentation map). However, their optimization tends to maximize the similarity to the original image while maintaining the ability to perform meaningful editing on local regions. This kind of entire synthesis can be easily identified if seen in training data~\cite{corvi2022detection}. To create more challenging fake media, we utilize the DALL-E2 model with high-quality local image editing techniques, which can generate text-guided pixels only in erased image regions.

\begin{table*}[t]
\renewcommand\arraystretch{1}
\centering
\caption{Summary of previous image manipulation datasets and our work.}
\vspace{-0.2cm}
\newcommand{\tabincell}[2]{\begin{tabular}{@{}#1@{}}#2\end{tabular}} 
% \vspace{-0.5mm}
\scalebox{0.85}{
\begin{tabular}{ l | c | c |c |c |c | c}
 \hline
{Dataset} & Year & \tabincell{c}{\# Forged Image} & \tabincell{c}{\# Authetic Image}  & Image Size & Format &\tabincell{c}{Manipulation Method} \\
\hline
Columbia~\cite{ng2004data} & 2004  & 912 & 933 & $128\times128$ & BMP & Random\\ \hline
Columbia~\cite{hsu06crfcheck} & 2006  & 180 & 183 & $757\times568$ - $1152\times768$ & TIF & Random\\ \hline
NIST16~\cite{NIST} & 2016 &  564 & 875 & $500\times500$ - $5616\times3744$& JPEG & Manual\\ \hline
CASIA v1~\cite{dong2013casia} & 2013  & 921  & 800 & $374\times256$ & JPEG & Manual\\ \hline
CASIA v2~\cite{dong2013casia} & 2013 & 5,123  & 7,200 & $320\times240$ - $800\times600$& JPEG, BMP, TIF & Manual\\ \hline
Coverage~\cite{wen2016coverage}& 2016 & 100 & 100 & $400\times486^*$ & TIF & Manual\\ \hline
\tabincell{l}{Realistic\\Tampering~\cite{korus2016evaluation}}& 2016 & 220 & 220 & $1920\times1080$ & TIF & Manual\\ \hline
%{FantasticReality~\cite{kniaz2019point}} & 2019 & 16,000 & 16,000  & $282\times800$ -$6000\times4000$ & JPEG & Manually\\ \hline
IMD2020~\cite{novozamsky2020imd2020}& 2020 & 2,010 & 414 & $1062\times866^*$ & JPEG, PNG & Collected from Internet \\ \hline
\textbf{AutoSplice (ours)}& \textbf{2023}  & \textbf{3,621} & \textbf{2,273}  & $\mathbf{256\times256}$ \textbf{-} $\mathbf{4232\times4232}$ & \textbf{JPEG} & \textbf{LLI model}\\
\hline
\end{tabular}}
\label{tab:t1}
\begin{tablenotes}%[flushleft]
\scriptsize
\item {* Using the average image size.}% provided by the original paper~\cite{wen2016coverage}.}
 \vspace{-0.3cm}
\end{tablenotes}
\end{table*}

\subsection{Image Forensic Datasets}
Two types of fake media datasets are relevant to our work: AI-synthesized image datasets and local image manipulation datasets.
Several large-scale AI-synthesized image datasets have been collected from various GAN and VAE models, including DFFD~\cite{dang2020detection} with GAN-based face attribute manipulations and entire face synthesis, CNNDetection~\cite{wang2020cnn} created using 11 CNN-based image generators, $DF^3$\cite{ju2022glff} with entire face generation from six generation models (i.e., StyleGAN2\cite{karras2020analyzing}, StyleGAN3~\cite{karras2021alias}, 3DGAN~\cite{chan2022efficient}, Taming Transformers~\cite{esser2021taming}, LSGM~\cite{vahdat2021score}, and Stable Diffusion~\cite{rombach2022high}), and DMimageDetection~\cite{corvi2022detection} with diverse images from different GAN and diffusion models. The recent diffusion-based generation models are a class of likelihood-based models~\cite{nichol2021improved}, which perturb data through successive addition of Gaussian noise and learn to recover the data by reversing this noising process. Although diffusion-based models achieve superior generation quality to GAN models~\cite{muller2022diffusion, stypulkowski2023diffused}, a study~\cite{corvi2022detection} showed that current GAN image detectors can achieve near-perfect detection on similar diffusion models when trained with images generated with the diffusion models.
%synthetic images from diffusion models can be identified with a high AUC with similar diffusion models in training~\cite{corvi2022detection}.

For local image manipulation, current research primarily focuses on techniques such as image slicing and copy-move. These methods involve copying and pasting specific regions of an image onto another part of the same image or a different one. Several widely used datasets, including Columbia~\cite{ng2004data, hsu06crfcheck}, CASIA~\cite{dong2013casia}, NIST16~\cite{NIST}, Coverage~\cite{wen2016coverage}, Realistic Tampering~\cite{korus2016evaluation}, and IMD2020~\cite{novozamsky2020imd2020}, offer various types of locally manipulated images that are created either by manual operations or random slicing. However, these datasets have certain limitations, such as small data size (e.g., Columbia, Coverage, and NIST16), low authenticity level (e.g., Columbia's random region copy-move), or low flexibility and efficiency in generation due to careful and manual operations (e.g., CASIA and Realistic Tampering).

Recent advances in large-scale language-image models have demonstrated remarkable abilities in text-guided image manipulation and generation. Leveraging the power of these models, we introduce AutoSplice, a text-prompt guided image manipulation dataset that is built using the DALL-E2 model for automatic image editing. Table~\ref{tab:t1} provides further details regarding the existing image local manipulation datasets and our AutoSplice dataset.

%In terms of caption-image inconsistency detection, one of the earliest datasets was proposed in 2017, named MAIM~\cite{jaiswal2017multimedia}, which comprises more than 239,000 pairs of randomly mismatched images and captions. The work~\cite{sabir2018deep} proposes the Multimodal Entity Image Repurposing (MEIR) dataset by swapping associated captions with randomly chosen captions, aiming to potentially result in semantically inconsistent repurposing. Similarly, the COSMOS dataset~\cite{aneja2021cosmos} uses random-chosen text to generate large-scale mismatched image-caption pairs in 2021. Instead of generating inconsistency samples based on random swapping, the NewsCLIPpings~\cite{luo2021newsclippings} dataset employs automatic retrieval to swap similar captions, and creates 988,000 image-caption pairs with the powerful multimodal model CLIP~\cite{CLIP_radford2021}. One of the challenges in these datasets is that the definition of inconsistency and consistency itself is obscure and can often be a moving target. For example, the study in \cite{huang2022text} has revealed that some inconsistency image and text pairs actually present a very consistent picture, therefore making automatic inconsistency detection methods hard to provide reliable predictions.

%\begin{figure}[t]
 %   \centering
%    \includegraphics[width=0.47 \textwidth]{Fig4.png}
%    \caption{\small Comparison of our dataset with existing local image manipulation datasets.}
%    \vspace{-0.35cm}
%    \label{fig:f4}
%\end{figure}
%\vspace{0.05cm}
\subsection{Image Forgery Detection}
The advancement of image forgery detection methods is a critical step in identifying manipulated/synthetic images for media forensics. Deep learning techniques have become increasingly popular for designing effective image forgery detectors. Most approaches consider forgery detection as a binary classification task and utilize well-designed deep neural networks to learn discriminative features automatically. Studies in this area can be divided into two categories: image-level forgery detection and pixel-level forgery detection (i.e., localization).

The former category concentrates on extracting global artifacts that synthesis models leave on the entire image, such as using augmented CNN models~\cite{wang2020cnn}, frequency analysis~\cite{frank2020leveraging}, and re-synthesis residuals~\cite{yang_ijcai21}. To improve the generalization ability for local image manipulation detection, recent studies~\cite{ju2022fusing, das2022gca, ju2022glff, lin2023image, zhuang2022uia} demonstrate the effectiveness of fusing local and global features in detecting different types of image forgery.

\begin{figure*}[t]
    \centering
    \includegraphics[width=1.0\textwidth]{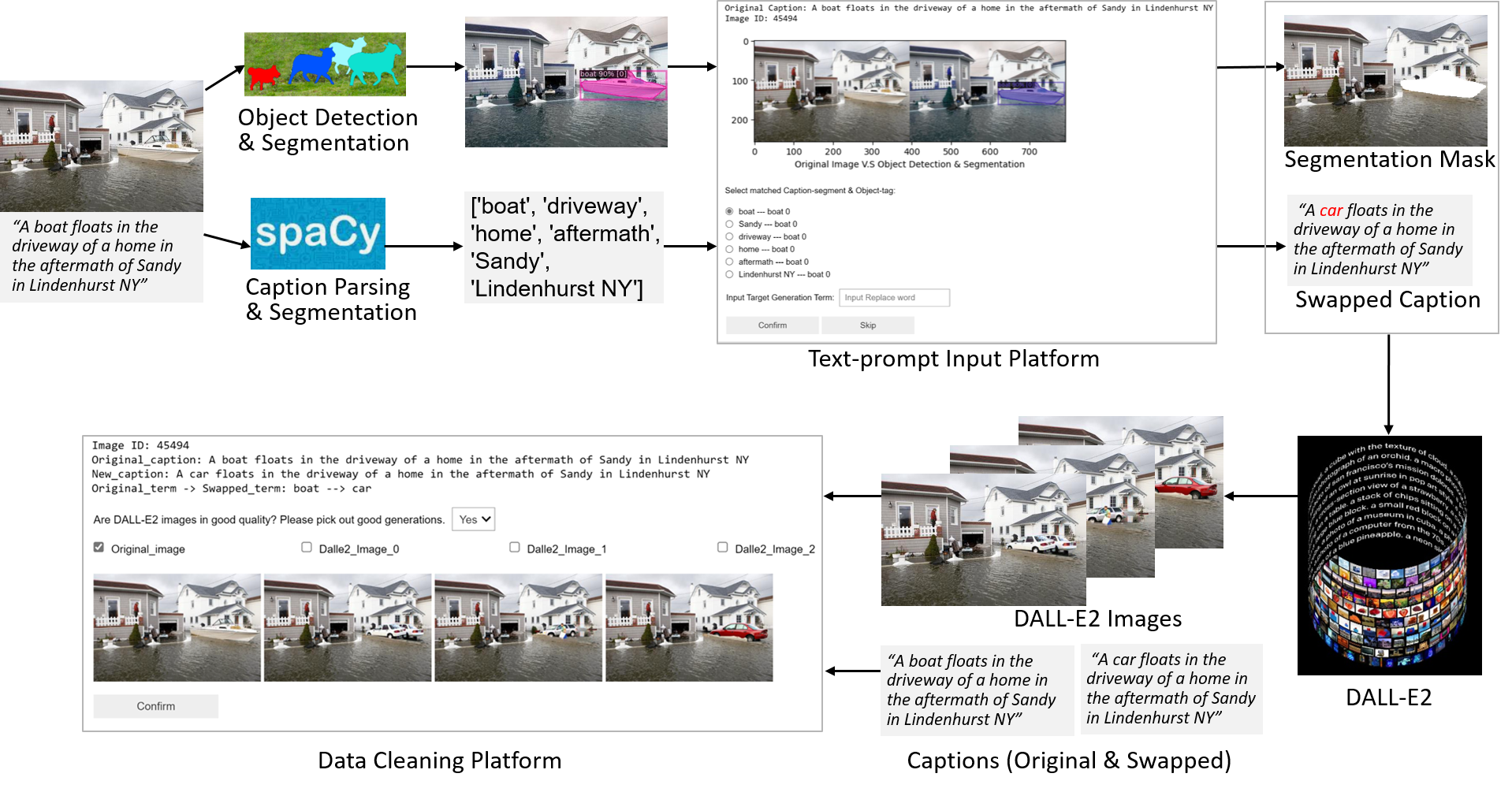}
    \vspace{-0.65cm}
    \caption{\small Pipeline of our text-prompt-based manipulation and annotation.}
    \vspace{-0.35cm}
    \label{fig:f2}
\end{figure*}

For manipulation localization, which aims to identify modifed image regions at the pixel level, existing methods mainly focus on identifying image tampering involving copy-move, splicing, removal, and faceswap. A subset of techniques formulates this task as a local anomaly detection problem and designs methods for capturing anomalies~\cite{cozzolino2016single, 2019ManTraNet, 2023ViTVAE}. Several methods~\cite{bianchi2012image,liu2018deep, 2021cat-v1,2022cat-v2, mareen2022comprint} utilize compression artifacts for forgery detection considering that the manipulation often
involves double or more times of compressions. In addition, a branch of methods explores distinctive noise patterns in forged images, such as the RGB-N model~\cite{zhou2018learning} fusing RGB image content and
image noise features, Noiseprint model with Siamese Network~\cite{2019noiseprint}, \cite{kong2022detect}, and ViT-VAE~\cite{2023ViTVAE} to combine Noiseprint, High-pass filtering residuals, and Laplacian edge maps using Vision Transformer (ViT). 

Given that the latest LLI models offer a powerful tool for realistic image generation and manipulation, presenting a potential threat of their misuse for spreading disinformation, our goal is to investigate the performance of current detectors on the newly created synthetic images. We will also pay particular attention to the detectors' ability in challenging social-network scenarios with image resizing and compression.

%-------------------------------------------------------
\vspace{-0.15cm}
\section{AutoSplice Dataset}
To evaluate the difficulty of detecting media generated by recent LLI models using current forensic techniques, we introduce a new dataset called AutoSplice. In this section, we provide a comprehensive overview of the AutoSplice dataset creation process. The entire generation pipeline is illustrated in Figure \ref{fig:f2}. We begin by outlining the data preprocessing techniques and human annotation process. Next, we describe the data cleaning process, followed by a summary of the dataset and an analysis of its statistics.

\vspace{-0.05cm}
\subsection{DALL-E2}
DALL-E2~\cite{DALLE2ramesh2022hiera} is a generative model that uses multiple modes to create synthetic images based on given text inputs. The model utilizes diffusion models to produce images based on CLIP image embeddings. Unlike traditional image synthesis approaches, DALL-E2 can also perform image inpainting by using both the input text and a region mask.

\subsection{Dataset Construction}

In this section, we introduce the details of the construction of the AutoSplice dataset. 

\vspace{-0.3cm}
\subsubsection{Pre-processing}
The pre-processing step for the DALL-E2 model involves providing a manipulation mask and a contextual text description to perform local image editing. In order to generate realistic manipulations, we propose using a specific object region with a modified caption where the corresponding text-prompt to the object is replaced with a target generation term. To achieve this, we use an object detection model to extract a list of object regions and a text parsing tool to segment text terms. We compare these terms with corresponding object-term pairs to facilitate further replacement and manipulation. Specifically, we use the Visual News dataset~\cite{liu2020visual}, which contains over one million news images along with their corresponding captions and metadata obtained from reputable real-world news sources (The Guardian, BBC, USA TODAY, and The Washington Post). This dataset has been utilized for various media forensic tasks, including the media manipulation detection~\cite{shao2023dgm4} and text-image inconsistency detection~\cite{huang2022text}. For each sample in Visual News, we utilize the Detic model~\cite{zhou2022detecting} to extract and segment object regions with detected object tags in the image. We also use spaCy \footnote{ \href{https://github.com/explosion/spaCy}{\url{https://github.com/explosion/spaCy}.}} for sentence segmentation and noun term extraction, as shown in Figure~\ref{fig:f2}. Human annotations are then used to select the object with corresponding descriptions in the caption, and input target terms to replace the object description in the caption.

\vspace{-0.3cm}
\subsubsection{Human Annotations}
Five annotators who are graduate and undergraduate students with professional backgrounds and have a clear understanding of the data annotation task for DALL-E2 input, strictly follow the steps outlined in Figure~\ref{fig:f2} during the data annotation process. 
\begin{enumerate}
\item Select the matching object region tag in the real-world image and the corresponding text description term in the caption (if present).
\vspace{-0.3cm}
\item Provide a target generation term that is similar but inconsistent with the original term and image.
\vspace{-0.3cm}
\item Ensure that the modified caption has the correct syntax.
\end{enumerate}
For each caption-image sample with a matched object-term pair, human annotations provide the two required inputs for the DALL-E2 model to perform local image generation: the segmented object region as the erased manipulation mask and the modified caption as the text prompt. The DALL-E2 model returns a group of three manipulation outputs for each generation.

\vspace{-0.3cm}
\subsubsection{Post-processing}

To address the limitations of the DALL-E2 model in generating human, text, and abstract concepts \cite{Imagen_saharia2022phot}, we conducted manual data cleaning to filter out images with visible visual artifacts or caption-image pairs with undesirable consistencies. Given an original image-caption pair with the swapped caption and three DALL-E2 generated images, annotators were required to assess the visual quality of each generated image and identify images with good quality (i.e., no obvious artifacts). Since the definition of ``good quality" is subjective and may vary among different annotators, each image was assessed multiple times by different annotators. We only retained images that received consistent labels from at least three annotators.

%We assigned a label consisting of three values to each sample 
The high-quality DALL-E2 images were resized to match the dimensions of their corresponding authentic images. Despite being initially in PNG format, we compressed the generated DALL-E2 images using JPEG for two reasons. Firstly, their corresponding authentic images included in our dataset are in JPEG format. It is essential to eliminate format-level clues in the binary image forgery classification task. Secondly, JPEG is the most important and widely used image compression format~\cite{sun2022lossless}, particularly on social media and websites, due to its simplicity and efficiency. Therefore, we chose both lossless (with a JPEG quality factor of 100) and gently lossy compression (with a quality factor of 90) formats to produce two variations of our DALL-E2 dataset.
%\vspace{-0.2cm}
%\begin{enumerate}
%  \item How many DALL-E2 generated images are there with good visual quality?
%\vspace{-0.2cm}
%  \item Is the original caption inconsistent with the DALL-E2 image? 
%\vspace{-0.2cm}
%  \item Is the swapped caption inconsistent with the original image?
%\end{enumerate}
%\vspace{-0.2cm}

\subsection{Dataset Summary}
\begin{figure}[t]
    \centering
    \includegraphics[width=0.5 \textwidth]{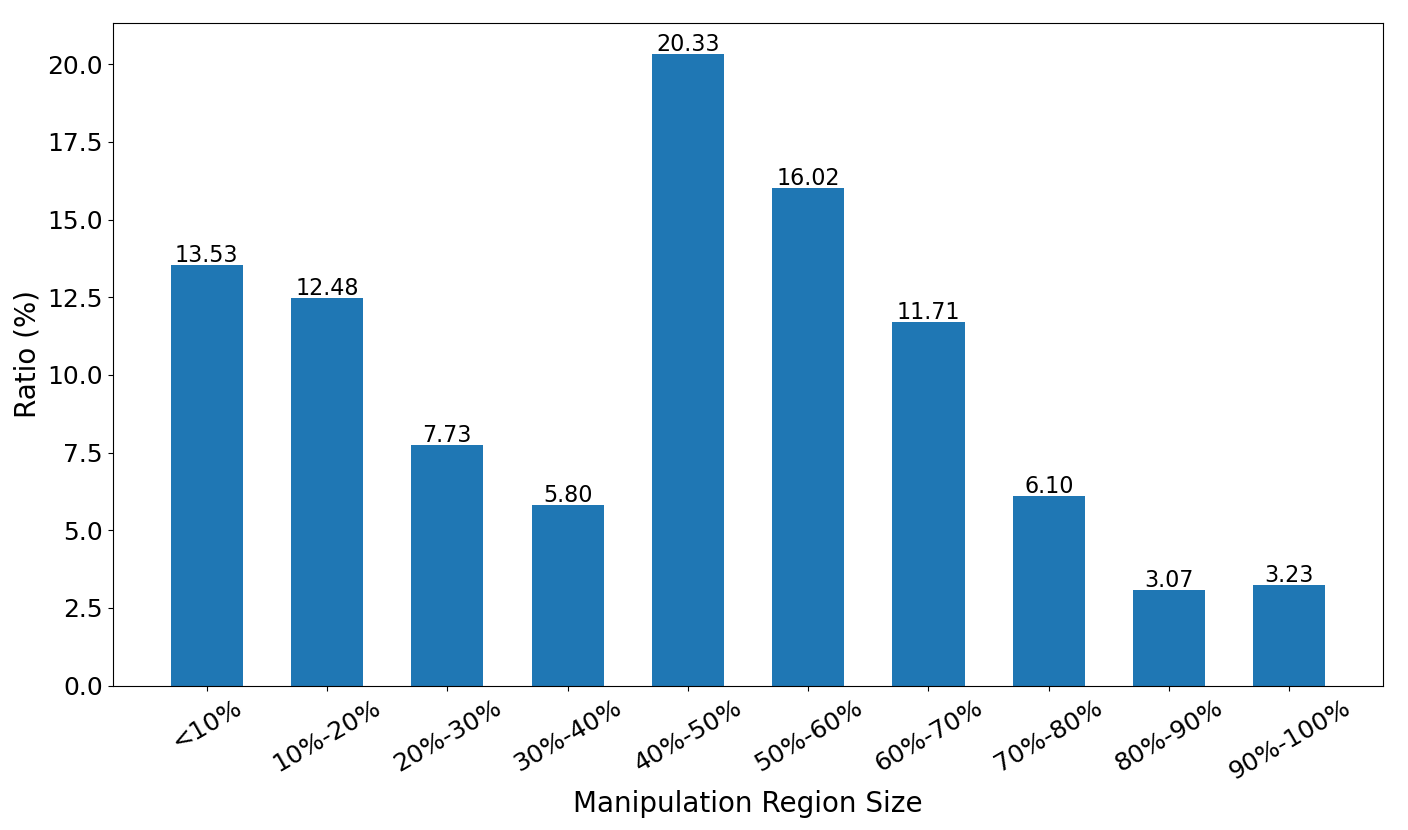}
    \vspace{-0.65cm}
    \caption{\small Statistical distribution of AutoSplice dataset in manipulation region size.}
    \vspace{-0.15cm}
    \label{fig:f3}
\end{figure}

\begin{figure}[h]
    \centering
    \includegraphics[width=0.48\textwidth]{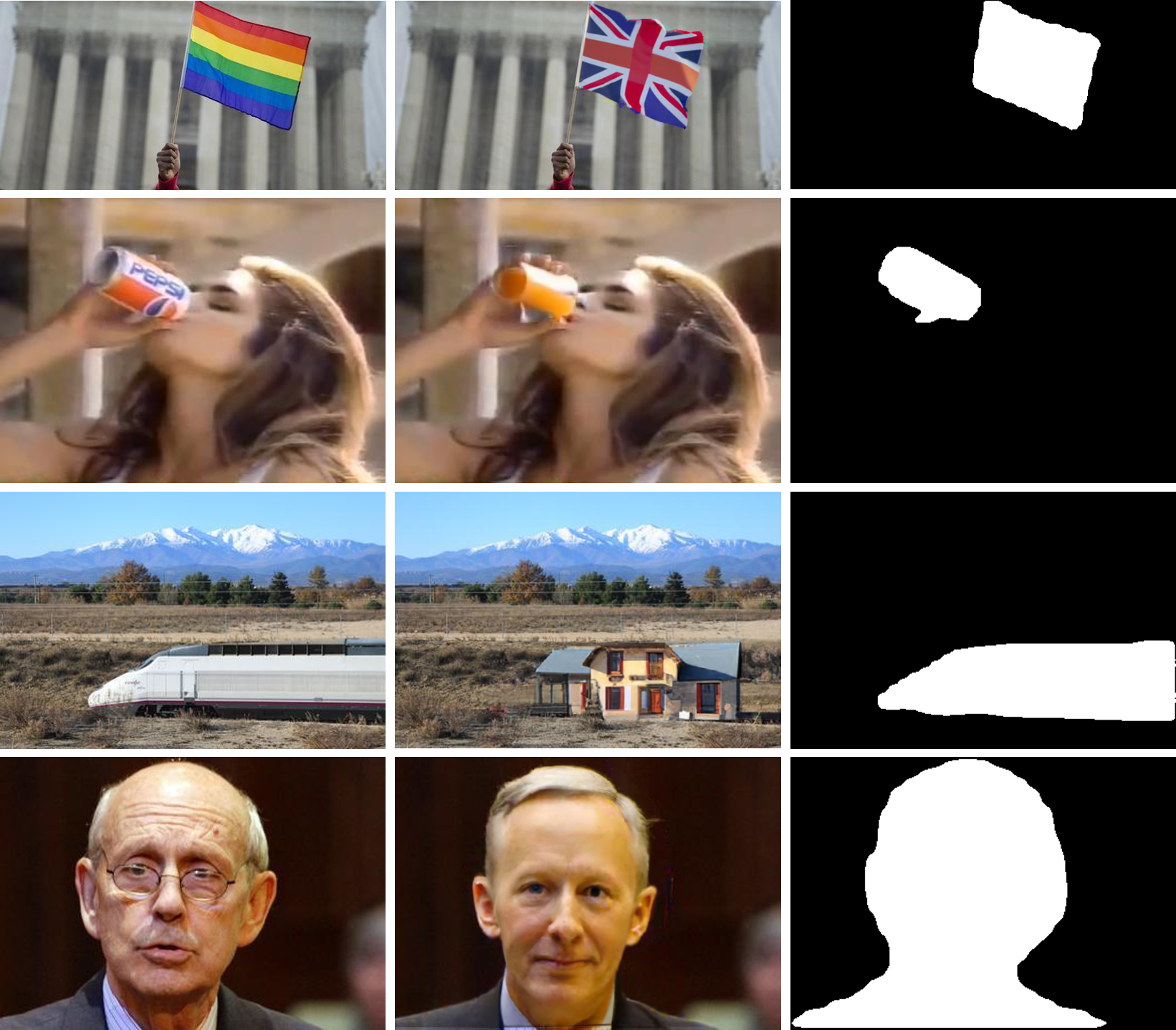}
    \vspace{-0.62cm}
    \caption{\small Examples in our AutoSplice dataset. The first column shows authentic images, while the second column displays forged images. The third column shows forgery masks.}
    \vspace{-0.35cm}
    \label{fig:f3a}
\end{figure}

Our AutoSplice dataset includes $3,621$ high-quality manipulated images and $2,273$ authentic images for each compression version. The data has been cleaned, and manipulation masks have been applied, allowing for further evaluation in both image forgery detection and image manipulation localization tasks. Figure \ref{fig:f3} presents statistical information about the size of the manipulation region within the dataset, indicating that the dataset has a high diversity in the manipulation region. We further show some image examples in Figure \ref{fig:f3a}.

%-------------------------------------------------------
\section{Experimental Evaluation}
This section outlines the evaluation experiments conducted on the AutoSplice dataset using state-of-the-art image-level and pixel-level forgery detection methods. The first experiment examines the generalization ability of existing pre-trained detectors to the LLI model manipulated images. % using pre-trained models. 
Following this, we analyze the performance limits of these detectors in in-domain testing scenarios, where the models are fine-tuned on our AutoSplice dataset for both detection and localization tasks.

\begin{table*}[h]
\newcommand{\tabincell}[2]{\begin{tabular}{@{}#1@{}}#2\end{tabular}} %
\centering
\caption{Image forgery detection and localization baselines}
\vspace{-0.2cm}
\scalebox{0.65}{
\begin{tabular}{l|c|c|c|c} 
\hline
Reference & Year &  Feature & Training set & Software Code \\ 
\hline
CNN-aug~\cite{wang2020cnn} & 2020 & Augmented CNN features & ProGAN~\cite{wang2020cnn} (720K images) &  \url{https://github.com/PeterWang512/CNNDetection}\\
\hline
\tabincell{l}{ResNet50\\ Nodown~\cite{gragnaniello2021gan}}  &  2021 & \tabincell{l}{No down-sampling\\CNN features} & ProGAN~\cite{wang2020cnn} (720K images) & \url{https://github.com/grip-unina/GANimageDetection} \\
\hline
\tabincell{l}{Beyondthe-\\Spectrum~\cite{yang_ijcai21}}  &  2021 & Re-synthesis residuals & ProGAN~\cite{wang2020cnn} (720K images) & \url{https://github.com/SSAW14/BeyondtheSpectrum} \\
\hline
PSM~\cite{ju2022fusing}  & 2022 & Global \& local features  & ProGAN~\cite{wang2020cnn} (720K images) & \url{https://github.com/littlejuyan/FusingGlobalandLocal}   \\
\hline
GLFF~\cite{ju2022glff}   &  2022 &   Multi-scale features  & ProGAN~\cite{wang2020cnn} (720K images) & \url{https://github.com/littlejuyan/GLFF}   \\  
\hline\hline
\hline
Noiseprint~\cite{2019noiseprint}  & 2019 & Noise residuals & 4 datasets with 125 cameras &  \url{https://grip-unina.github.io/noiseprint/}  \\
\hline
ManTra-Net~\cite{2019ManTraNet}& 2019 & Anomalous features &  4 synthetic datasets  &     \url{https://github.com/ISICV/ManTraNet}   \\
\hline 
ForensicsGraph~\cite{2020forensicgraphs} & 2020   &  Similarity graph  &  4 million image patches from 80 cameras   &   \url{https://gitlab.com/omayer/forensic-graph}   \\
\hline
CAT-Net~\cite{2021cat-v1} & 2021   &  Compression artifacts  &   4 synthetic datasets (960K images)    & \url{https://github.com/mjkwon2021/CAT-Net}  \\
\hline
%CAT-Net-v2~\cite{2022cat-v2} & 2022   &  Compression artifacts  &  5 custom datasets   &   \url{https://github.com/mjkwon2021/CAT-Net}     \\
%\hline
MVSS-Net~\cite{2021MVSSNet}& 2021   &  Multi-view features      & 1 dataset (CASIA v2)   &   \url{https://github.com/dong03/MVSS-Net}   \\
\hline
PSCC-Net~\cite{liu2022pscc} & 2022   &   Spatio-channel correlation    &  0.38 million images    & \url{https://github.com/proteus1991/PSCC-Net}  \\  
\hline
ViT-VAE~\cite{2023ViTVAE} & 2023   &  Multi-modal features    &  $-^*$   &   \url{https://github.com/media-sec-lab/ViT-VAE}  \\
\hline             
\end{tabular} } 
\begin{tablenotes}
\scriptsize
\item {* Using run-time training where the ViT-VAE model requires an independent training phase for each test image.}
 \vspace{-0.35cm}
\end{tablenotes}
\label{tab:2}
\end{table*}

\subsection{Evaluation Baselines}
%\noindent \textbf{Image Forgery Detection.}
We evaluated five image forgery detection methods on our dataset: CNN-aug~\cite{wang2020cnn}, ResNet50 Nodown~\cite{gragnaniello2021gan}, BeyondtheSpectrum~\cite{yang_ijcai21}, PSM~\cite{ju2022fusing}, and GLFF~\cite{ju2022glff}. These models were chosen because they have demonstrated excellent performance in detecting image forgery, have been evaluated on both globally and locally manipulated images, and provide open-source codes and pre-trained models.

For pixel-level forgery detection and localization, we evaluated seven techniques that employ different feature learning strategies: Noiseprint~\cite{2019noiseprint}, ManTra-Net~\cite{2019ManTraNet}, ForensicsGraph~\cite{2020forensicgraphs}, CAT-Net~\cite{2021cat-v1}, MVSS-Net~\cite{2021MVSSNet}, PSCC-Net~\cite{liu2022pscc}, and ViT-VAE~\cite{2023ViTVAE}. Table~\ref{tab:2} presents detailed information on the feature design, training set, and software codes of these baseline methods.

\subsection{Evaluation Metrics}

To detect binary image forgery, we assign Positive (1) to the tampered image/pixel and Negative (0) to the authentic image/pixel. In image-level detection, we use standard metrics such as Area Under ROC Curve (AUC), True Positive Rate (TPR), and False Positive Rate (FPR). For pixel-level detection, we calculate F1 score, Intersection over Union (IoU), precision, and accuracy (ACC) by comparing the results with the binary ground-truth mask using a fixed threshold of 0.5, as done in previous studies~\cite{2023ViTVAE, dong2022mvss, 2021MVSSNet, chen2021image, wang2022jpeg}. The average scores for all testing images are reported.

\noindent{\bf Experiment Settings}.
To evaluate our dataset, we adhered to the parameter settings used in the baseline implementation for each task. During the fine-tuning process, we set a 6:4 split between training and testing data to ensure that there was no overlapping between the two sets. 

\subsection{Comparisons on Image Forgery Detection}
\subsubsection{Pre-trained Models} 
We started by using five pre-trained models for image forgery detection and tested their ability to identify the AutoSplice forgery. The results, including AUC, TPR, and FPR, are presented in Table \ref{tab:3}. Our analysis shows that all models trained on the ProGAN dataset \cite{wang2020cnn}, which includes 720K images (360K real images and 360K fake images across 20 object categories), experienced a performance decrease in detecting image forgery in the AutoSplice dataset. Only the ResNet Nodown\cite{gragnaniello2021gan} model achieved the best AUC on two compression sets. All other models had an AUC lower than 0.600. The low TPR and FPR scores suggest that most LLI model forged images were incorrectly classified as authentic, whereas authentic images were correctly identified. Additionally, all models performed poorly on JPEG-90 images with mild compression, indicating that the compression process reduces the distinguishability of features.

\subsubsection{Fine-tuned Models}
\label{4.3.2}
We fine-tuned four models with training codes on the AutoSplice training dataset, after considering that data-driven classification methods tend to perform better when the domain discrepancy is alleviated~\cite{liu2022pscc}. To ensure that compression artifacts did not influence the binary forgery detection task, we compressed the original images produced by the DALL-E2 model using the same JPEG compression quality factor (75 \footnote{The JPEG-75 compressed images are included in our AutoSplice dataset along with JPEG-100 and JPEG-90 versions at \url{https://github.com/shanface33/AutoSplice_Dataset.}}) as the authentic images derived from Visual News dataset~\cite{liu2020visual}. We evaluated the detection performance on two testing subsets (JPEG-100 and JPEG-90) and reported the results in Table \ref{tab:4}. As expected, most methods showed a significant improvement in the detection AUC and TPR when evaluated in the in-domain testing scenario. The CNN-aug~\cite{wang2020cnn} achieved the best performance on both compression sets. However, after fine-tuning, most methods had a significant increase in FPR, indicating that a greater number of authentic images were incorrectly classified.

\begin{table}[h]
\newcommand{\tabincell}[2]{\begin{tabular}{@{}#1@{}}#2\end{tabular}} %
\centering
\caption{Image forgery detection results on AutoSplice dataset using pre-trained models. Best results are shown in bold.}
\vspace{-0.2cm}
\scalebox{0.83}{
\begin{tabular}{l|c|c|c|c|c|c} 
\hline
\multirow{2}{*}{Method}  & \multicolumn{3}{c|}{JPEG - 100}  & \multicolumn{3}{c}{JPEG - 90} \\
\cline{2-7}
&  {AUC\textuparrow } & {TPR\textuparrow } & {FPR\textdownarrow } & {AUC\textuparrow } & {TPR\textuparrow } & {FPR\textdownarrow }  \\
\hline
CNN-aug~\cite{wang2020cnn} & 0.597 & 0.025 & 0.004 & 0.551 & 0.004 & 0.004    \\
\hline
\tabincell{l}{ResNet50\\ Nodown~\cite{gragnaniello2021gan}}  & \textbf{0.750} & 0.070 & \textbf{0.002} & \textbf{0.664} & 0.004 & \textbf{0.002}  \\
\hline
\tabincell{l}{Beyondthe-\\Spectrum~\cite{yang_ijcai21}}  & 0.547 & \textbf{0.335} & 0.290 & 0.503 & \textbf{0.303} & 0.290  \\
\hline
PSM~\cite{ju2022fusing}  &  0.586 & 0.038 & \textbf{0.002} & 0.535 & 0.005  & \textbf{0.002} \\
\hline
GLFF~\cite{ju2022glff} & 0.572 & 0.055 & 0.013 & 0.526 & 0.016 & 0.013           \\   
\hline        
\end{tabular} } 
\vspace{-0.45cm}
\label{tab:3}
\end{table}

\begin{table} [h]
\newcommand{\tabincell}[2]{\begin{tabular}{@{}#1@{}}#2\end{tabular}} %
\centering
\caption{Image forgery detection results on AutoSplice dataset using fine-tuned models. Best results are shown in bold.}
\vspace{-0.2cm}
\scalebox{0.83}{
\begin{tabular}{l|c|c|c|c|c|c}
\hline
\multirow{2}{*}{Method}   & \multicolumn{3}{c|}{JPEG - 100}    & \multicolumn{3}{c}{JPEG - 90}\\
\cline{2-7} & {AUC\textuparrow } & {TPR\textuparrow } & {FPR\textdownarrow } & {AUC\textuparrow } & {TPR\textuparrow } & {FPR\textdownarrow }   \\ \hline
 CNN-aug~\cite{wang2020cnn} & \textbf{0.979} & \textbf{0.981} & 0.372 & \textbf{0.948} & \textbf{0.932} & 0.372 \\
 %0.976 & 0.939 & 0.090  & 0.949 & 0.843 & 0.090 \\
\hline
\tabincell{l}{Beyondthe-\\Spectrum~\cite{yang_ijcai21}}   & 0.797 & 0.785 & 0.341 & 0.787 & 0.762 & 0.341\\
%0.937 & 0.952 & 0.319 & 0.796 & 0.776 & 0.319 \\
\hline
PSM~\cite{ju2022fusing}& 0.880 & 0.847 & 0.257 & 0.882 & 0.841 & 0.257 \\
%0.949 & 0.888 & 0.129 & 0.863 & 0.694 & 0.129\\
\hline
GLFF~\cite{ju2022glff} & 0.926 & 0.776 & \textbf{0.077} & 0.908 & 0.735 & \textbf{0.077} \\
%0.963 & 0.921 & 0.127 & 0.891 & 0.768 & 0.127 \\ 
\hline                             
\end{tabular}} 
\vspace{-0.15cm}
\label{tab:4}
\end{table}

\begin{table*}[t]
\newcommand{\tabincell}[2]{\begin{tabular}{@{}#1@{}}#2\end{tabular}} %
\centering
\caption{Image forgery localization results on AutoSplice dataset using pre-trained models. Best results are shown in bold.}
\vspace{-0.2cm}
\scalebox{0.95}{
\begin{tabular}{l|c|c|c|c|c|c|c|c|c} 
\hline
\multirow{2}{*}{Method} &\multicolumn{4}{c|}{Forged JPEG-100} & \multicolumn{4}{c|}{Forged JPEG-90} & {Authentic} \\
\cline{2-10}
&  F1\textuparrow  & IoU\textuparrow  & Precision\textuparrow & ACC\textuparrow &F1\textuparrow  & IoU\textuparrow  & Precision\textuparrow & ACC\textuparrow  & ACC\textuparrow\\ 
\hline
Noiseprint~\cite{2019noiseprint} & 0.333 & 0.217 & 0.390 & 0.480 & 0.316 & 0.205 & 0.373 & 0.467 &  0.594    \\
\hline
ManTra-Net~\cite{2019ManTraNet} & 0.179 & 0.120 & 0.639 & 0.586 & 0.062 & 0.035 & 0.586 & 0.716  &  0.992    \\
\hline 
ForensicsGraph~\cite{2020forensicgraphs} &0.362  & 0.289 &  0.393 & 0.530 & 0.354 & 0.253  & 0.438 & 0.518 &  0.584   \\
\hline
CAT-Net~\cite{2021cat-v1}& \textbf{0.751} & \textbf{0.648} & 0.884 & \textbf{0.827} & \textbf{0.676} & \textbf{0.579} & \textbf{0.833} & \textbf{0.793}  &  0.933  \\
\hline
MVSS-Net~\cite{2021MVSSNet} & 0.330 & 0.238 & 0.734 & 0.677 & 0.141 & 0.093 & 0.516 & 0.612  &  0.991   \\
\hline
PSCC-Net~\cite{liu2022pscc} & 0.558 & 0.447 & \textbf{0.899} & 0.725 & 0.056 & 0.036 & 0.295 & 0.591  & \textbf{0.998}   \\
%0.960 & 0.928 & 0.964 & 0.976 & 0.001 & 0.001 & 0.009 & 0.578   \\  
\hline
ViT-VAE~\cite{2023ViTVAE} & 0.156 & 0.115 & 0.245 & 0.560 & 0.244 & 0.183 & 0.275 & 0.534  & 0.835 \\  
\hline             
\end{tabular} } 
    \vspace{-0.10cm}
\label{tab:5}
\end{table*}

\begin{figure*}[t]
    \centering
    \includegraphics[width=0.95 \textwidth]{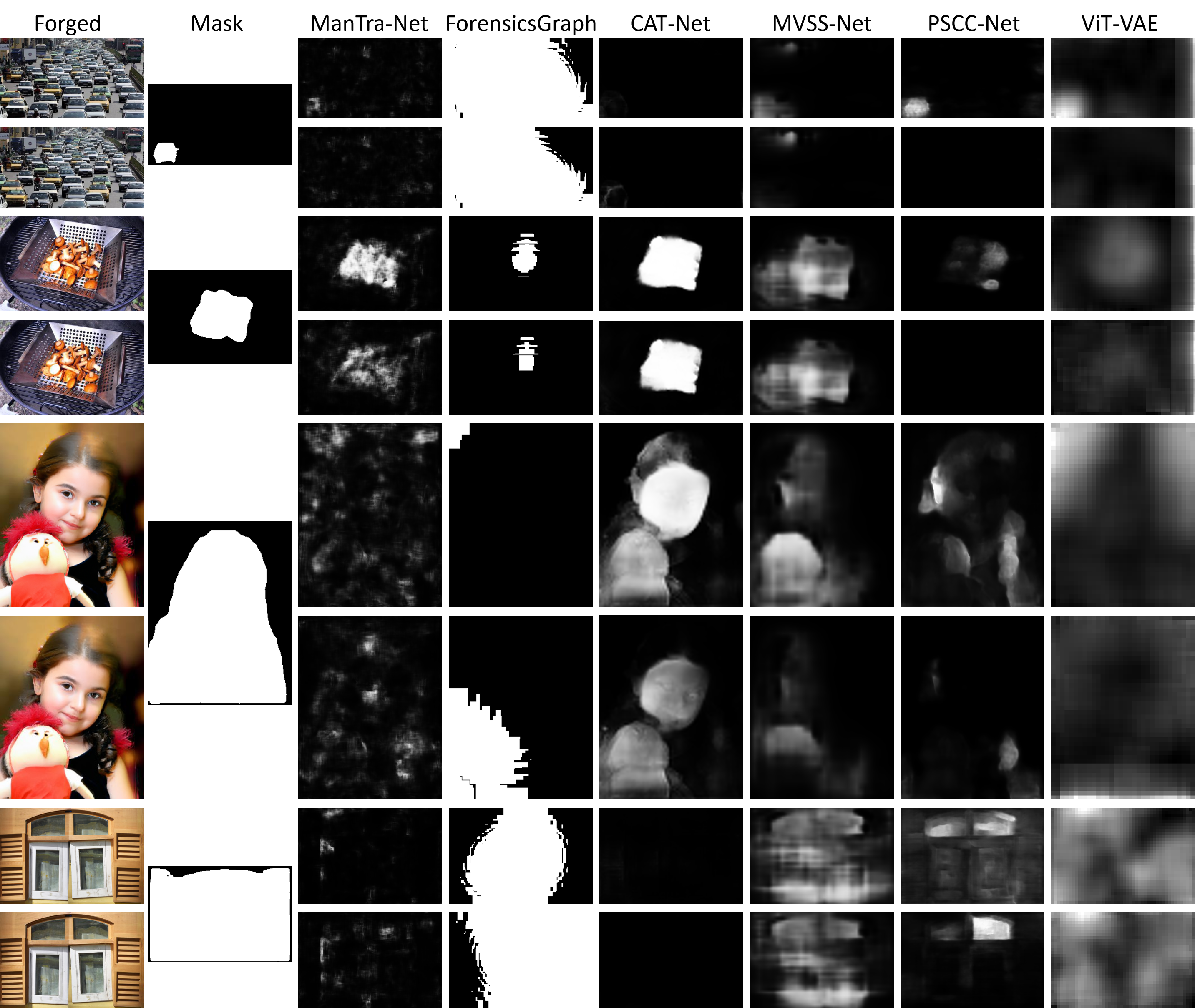}
    \caption{\small Localization results of pre-trained models in detecting AutoSplice forged images with different manipulation regions. The images in odd rows are compressed using JPEG-100, while the images in even rows are compressed using JPEG-90.}
    \vspace{-0.35cm}
    \label{fig:f4}
\end{figure*}

\begin{table*}
\newcommand{\tabincell}[2]{\begin{tabular}{@{}#1@{}}#2\end{tabular}} %
\centering
\caption{Image forgery localization results on AutoSplice dataset using fine-tuned models. Best results are shown in bold.}
\vspace{-0.2cm}
\scalebox{0.95}{
\begin{tabular}{l|c|c|c|c|c|c|c|c|c }
\hline
\multirow{2}{*}{Method}   & \multicolumn{4}{c|}{JPEG - 100}    & \multicolumn{4}{c|}{JPEG - 90} & {Authentic} \\
\cline{2-10} 
& F1\textuparrow &  IoU\textuparrow  & Precision\textuparrow & ACC\textuparrow  & F1\textuparrow &  IoU\textuparrow  & Precision\textuparrow & ACC\textuparrow  & ACC\textuparrow  \\ \hline
%CAT-Net~\cite{2021cat-v1}   &  0.950 & 0.908 & 0.919 & 0.971 & \textbf{0.007} & \textbf{0.003} &  \textbf{0.201} &  \textbf{0.593} &  \textbf{1.000}   \\
CAT-Net~\cite{2021cat-v1}   &  0.762  & 0.658 & \textbf{0.882} & 0.837 & 0.693 & 0.594 & \textbf{0.844} & 0.805 &  0.927\\
\hline 
%PSCC-Net~\cite{liu2022pscc}   &   \textbf{0.960} & \textbf{0.928} & \textbf{0.965} & \textbf{0.976}  & 0.001 & 0.000 & 0.009 & 0.577 &  \textbf{1.000}    \\  
PSCC-Net~\cite{liu2022pscc} & \textbf{0.862} & \textbf{0.794} & 0.847 & \textbf{0.919} & \textbf{0.771} & \textbf{0.693} & 0.775 & \textbf{0.886} & \textbf{0.993} \\
\hline             
\end{tabular} } 
 \vspace{-0.35cm}
\label{tab:6}
\end{table*}

\subsection{Comparisons on Image Forgery Localization}
\subsubsection{Pre-trained Models}

We evaluated seven baselines for image forgery localization, and reported their pixel-level metrics in Table \ref{tab:5}. Results varied significantly across models due to their different training data and forensics cues (detailed in Table~\ref{tab:2}). The CAT-Net~\cite{2021cat-v1} model performed the best thanks to its two-stream network that learns compression artifacts from both RGB and DCT domains, as well as its extensive and diverse training data. The PSCC-Net~\cite{liu2022pscc} also outperformed other methods on the JPEG-100 testing set, but its performance degraded significantly on the JPEG-90 set. Most models showed poor generalization ability on the AutoSplice dataset, with F1 and IoU lower than 0.37 and 0.29, respectively. For authentic images, where every pixel in the ground truth mask is negative, F1, IoU, and Precision metrics are not appropriate. We reported the ACC in Table \ref{tab:5}, and most methods achieved high accuracy in detecting authentic images derived from real-world media data.

We want to note that the optimal threshold for each localization method may not be exactly 0.5, and it varies for different models and images. To eliminate the influence of the threshold, we further compared several examples with the predicted masks of these models without binarizing the map. Figure \ref{fig:f4} shows the results on AutoSplice forged images with two compression versions and different manipulation regions. We observed the influence of even mild compression on the localization performance when comparing the results in the odd rows on JPEG-100 compressed images and even rows on JPEG-90 images. Moreover, the size of the manipulation region appeared to be another crucial factor affecting the accuracy of localization. The majority of models struggled to detect forgeries containing large tampered regions in the AutoSplice dataset.

\vspace{-0.2cm}
\subsubsection{Fine-tuned Models} 
We fine-tuned two localization methods, CAT-Net~\cite{2021cat-v1,2022cat-v2} and PSCC-Net~\cite{liu2022pscc}, on the AutoSplice training set using JPEG-75 compressed images (as previously described in Section \ref{4.3.2}) to evaluate their performance in localizing pixel-level forgery. As expected, both models outperformed pre-trained models on the JPEG-100 and JPEG-90 testing sets, with the PSCC-Net model demonstrating significant improvement. However, the performance on the JPEG-90 set decreases obviously in comparison to the results on the JPEG-100 set, which aligns with the findings from Table \ref{tab:5}.
%As expected, the models demonstrated improvement in performance on the JPEG-100 testing set and achieved perfect accuracy on authentic sets during in-domain testing. However, both models failed to detect forgery with JPEG-90 compressed data. This may be attributed to the models overfitting the training set, leading to poor generalization ability.

%-------------------------------------------------------
\section{Conclusions}

This paper investigates the challenge posed by language-image generation models to media forensics and proposes a new approach that utilizes the DALL-E2 language-image model to splice masked regions guided by a text prompt. To ensure the creation of realistic manipulations, we have developed an annotation platform with human verification to validate reasonable text prompts. The approach has resulted in the creation of a new image dataset called {\em AutoSplice}, containing 5,894 manipulated and authentic images, including $3,621$ images generated by locally or globally manipulating real-world image-caption pairs, which we believe will be a valuable resource for future research. We have evaluated the effectiveness of several state-of-the-art forgery detectors in various testing scenarios. However, our experiments with pre-trained models revealed unsatisfactory generalization performance in forgery detection and localization. Including our dataset in training could enhance the performance of existing models during in-domain testing. This finding emphasizes that fine-tuning on datasets with homogeneous characteristics results in significant performance improvements in media forensics. Nevertheless, achieving balanced performance across different JPEG compression quality factors and tampered region sizes remains a challenging task for forgery localization.

For future works, we will consider: first, exploring more advanced models to generate more realistic manipulated images to further challenge media forensics. Secondly, investigating approaches to improve the generalization performance of forgery detection and localization models to handle various types of image manipulations. Thirdly, conducting experiments to evaluate the performance of existing forgery detection and localization models on the proposed dataset under various testing scenarios and compare them with state-of-the-art approaches. Last but not least, exploring the potential of transfer learning approaches to enhance the performance of existing models on new datasets with limited training samples.

\section{Impact Statement}

The recent advancements in language-image models have led to highly realistic image generation from textual descriptions, which can pose a potential threat to media forensics. This paper provides a new image dataset called AutoSplice,  created using the DALL-E2 language-image model to splice masked regions guided by a text prompt. The unsatisfactory generalization performance of existing forgery detection and localization models on the proposed dataset highlights the need for further investigation and improvement in this area. Future works proposed in this paper, such as exploring more advanced models and transfer learning approaches, aim to address these challenges and contribute to the advancement of media forensics research. Ultimately, this work will help detect image manipulations in various applications, including social media, journalism, and law enforcement, and contribute to ensuring the authenticity and reliability of digital media content.
%-------------------------------------------------------
\paragraph{Acknowledgement.} This work is supported by the US Defense Advanced Research Projects Agency (DARPA) Semantic Forensic (SemaFor) program, under Contract No. HR001120C0123. %The views, opinions and/or findings expressed are those of the authors and should not be interpreted as representing the official views or policies of the Department of Defense or the U.S. Government.

\clearpage
%%%%%%%%% REFERENCES
{\small
\bibliographystyle{ieee_fullname}
\bibliography{egbib}

\begin{thebibliography}{10}\itemsep=-1pt

\bibitem{NIST}
Nist nimble 2016 datasets., 2016.
\newblock Accessed: 2023-03-11.

\bibitem{bianchi2012image}
Tiziano Bianchi and Alessandro Piva.
\newblock Image forgery localization via block-grained analysis of jpeg
  artifacts.
\newblock {\em IEEE Transactions on Information Forensics and Security},
  7(3):1003--1017, 2012.

\bibitem{brock2018large}
Andrew Brock, Jeff Donahue, and Karen Simonyan.
\newblock Large scale gan training for high fidelity natural image synthesis.
\newblock {\em arXiv preprint arXiv:1809.11096}, 2018.

\bibitem{chan2022efficient}
Eric~R Chan, Connor~Z Lin, Matthew~A Chan, Koki Nagano, Boxiao Pan, Shalini
  De~Mello, Orazio Gallo, Leonidas~J Guibas, Jonathan Tremblay, Sameh Khamis,
  et~al.
\newblock Efficient geometry-aware 3d generative adversarial networks.
\newblock In {\em Proceedings of the IEEE/CVF Conference on Computer Vision and
  Pattern Recognition}, pages 16123--16133, 2022.

\bibitem{chang2023muse}
Huiwen Chang, Han Zhang, Jarred Barber, AJ Maschinot, Jose Lezama, Lu Jiang,
  Ming-Hsuan Yang, Kevin Murphy, William~T Freeman, Michael Rubinstein, et~al.
\newblock Muse: Text-to-image generation via masked generative transformers.
\newblock {\em arXiv preprint arXiv:2301.00704}, 2023.

\bibitem{2023ViTVAE}
Tong Chen, Bin Li, and Jinhua Zeng.
\newblock Learning traces by yourself: Blind image forgery localization via
  anomaly detection with vit-vae.
\newblock {\em IEEE Signal Processing Letters}, 2023.

\bibitem{2021MVSSNet}
Xinru Chen, Chengbo Dong, Jiaqi Ji, Juan Cao, and Xirong Li.
\newblock Image manipulation detection by multi-view multi-scale supervision.
\newblock In {\em Proceedings of the IEEE/CVF International Conference on
  Computer Vision}, pages 14185--14193, 2021.

\bibitem{chen2021image}
Xinru Chen, Chengbo Dong, Jiaqi Ji, Juan Cao, and Xirong Li.
\newblock Image manipulation detection by multi-view multi-scale supervision.
\newblock In {\em Proceedings of the IEEE/CVF International Conference on
  Computer Vision}, pages 14185--14193, 2021.

\bibitem{corvi2022detection}
Riccardo Corvi, Davide Cozzolino, Giada Zingarini, Giovanni Poggi, Koki Nagano,
  and Luisa Verdoliva.
\newblock On the detection of synthetic images generated by diffusion models.
\newblock {\em arXiv preprint arXiv:2211.00680}, 2022.

\bibitem{couairon2022diffedit}
Guillaume Couairon, Jakob Verbeek, Holger Schwenk, and Matthieu Cord.
\newblock Diffedit: Diffusion-based semantic image editing with mask guidance.
\newblock {\em arXiv preprint arXiv:2210.11427}, 2022.

\bibitem{cozzolino2016single}
Davide Cozzolino and Luisa Verdoliva.
\newblock Single-image splicing localization through autoencoder-based anomaly
  detection.
\newblock In {\em 2016 IEEE international workshop on information forensics and
  security (WIFS)}, pages 1--6. IEEE, 2016.

\bibitem{2019noiseprint}
Davide Cozzolino and Luisa Verdoliva.
\newblock Noiseprint: A cnn-based camera model fingerprint.
\newblock {\em IEEE Transactions on Information Forensics and Security},
  15:144--159, 2019.

\bibitem{dang2020detection}
Hao Dang, Feng Liu, Joel Stehouwer, Xiaoming Liu, and Anil~K Jain.
\newblock On the detection of digital face manipulation.
\newblock In {\em Proceedings of the IEEE/CVF Conference on Computer Vision and
  Pattern recognition}, pages 5781--5790, 2020.

\bibitem{daniel2021soft}
Tal Daniel and Aviv Tamar.
\newblock Soft-introvae: Analyzing and improving the introspective variational
  autoencoder.
\newblock In {\em Proceedings of the IEEE/CVF Conference on Computer Vision and
  Pattern Recognition}, pages 4391--4400, 2021.

\bibitem{das2022gca}
Sowmen Das, Md Islam, Md Amin, et~al.
\newblock Gca-net: utilizing gated context attention for improving image
  forgery localization and detection.
\newblock In {\em Proceedings of the IEEE/CVF Conference on Computer Vision and
  Pattern Recognition}, pages 81--90, 2022.

\bibitem{dhariwal2021diffusion}
Prafulla Dhariwal and Alexander Nichol.
\newblock Diffusion models beat gans on image synthesis.
\newblock {\em Advances in Neural Information Processing Systems},
  34:8780--8794, 2021.

\bibitem{ding2021cogview}
Ming Ding, Zhuoyi Yang, Wenyi Hong, Wendi Zheng, Chang Zhou, Da Yin, Junyang
  Lin, Xu Zou, Zhou Shao, Hongxia Yang, et~al.
\newblock Cogview: Mastering text-to-image generation via transformers.
\newblock {\em Advances in Neural Information Processing Systems},
  34:19822--19835, 2021.

\bibitem{dong2022mvss}
Chengbo Dong, Xinru Chen, Ruohan Hu, Juan Cao, and Xirong Li.
\newblock Mvss-net: Multi-view multi-scale supervised networks for image
  manipulation detection.
\newblock {\em IEEE Transactions on Pattern Analysis and Machine Intelligence},
  2022.

\bibitem{dong2013casia}
Jing Dong, Wei Wang, and Tieniu Tan.
\newblock Casia image tampering detection evaluation database.
\newblock In {\em 2013 IEEE China summit and international conference on signal
  and information processing}, pages 422--426. IEEE, 2013.

\bibitem{esser2021taming}
Patrick Esser, Robin Rombach, and Bjorn Ommer.
\newblock Taming transformers for high-resolution image synthesis.
\newblock In {\em CVPR}, pages 12873--12883, 2021.

\bibitem{frank2020leveraging}
Joel Frank, Thorsten Eisenhofer, Lea Sch{\"o}nherr, Asja Fischer, Dorothea
  Kolossa, and Thorsten Holz.
\newblock Leveraging frequency analysis for deep fake image recognition.
\newblock In {\em International conference on machine learning}, pages
  3247--3258. PMLR, 2020.

\bibitem{gafni2022make}
Oran Gafni, Adam Polyak, Oron Ashual, Shelly Sheynin, Devi Parikh, and Yaniv
  Taigman.
\newblock Make-a-scene: Scene-based text-to-image generation with human priors.
\newblock In {\em Computer Vision--ECCV 2022: 17th European Conference, Tel
  Aviv, Israel, October 23--27, 2022, Proceedings, Part XV}, pages 89--106.
  Springer, 2022.

\bibitem{gragnaniello2021gan}
Diego Gragnaniello, Davide Cozzolino, Francesco Marra, Giovanni Poggi, and
  Luisa Verdoliva.
\newblock Are gan generated images easy to detect? a critical analysis of the
  state-of-the-art.
\newblock In {\em 2021 IEEE international conference on multimedia and expo
  (ICME)}, pages 1--6. IEEE, 2021.

\bibitem{yang_ijcai21}
Yang He, Ning Yu, Margret Keuper, and Mario Fritz.
\newblock Beyond the spectrum: Detecting deepfakes via re-synthesis.
\newblock In {\em 30th International Joint Conference on Artificial
  Intelligence (IJCAI)}, 2021.

\bibitem{hertz2022prompt}
Amir Hertz, Ron Mokady, Jay Tenenbaum, Kfir Aberman, Yael Pritch, and Daniel
  Cohen-Or.
\newblock Prompt-to-prompt image editing with cross attention control.
\newblock {\em arXiv preprint arXiv:2208.01626}, 2022.

\bibitem{hsu06crfcheck}
Y.-F. Hsu and S.-F. Chang.
\newblock Detecting image splicing using geometry invariants and camera
  characteristics consistency.
\newblock In {\em International Conference on Multimedia and Expo}, 2006.

\bibitem{huang2022text}
Mingzhen Huang, Shan Jia, Ming-Ching Chang, and Siwei Lyu.
\newblock Text-image de-contextualization detection using vision-language
  models.
\newblock In {\em ICASSP 2022-2022 IEEE International Conference on Acoustics,
  Speech and Signal Processing (ICASSP)}, pages 8967--8971. IEEE, 2022.

\bibitem{ju2022glff}
Yan Ju, Shan Jia, Jialing Cai, Haiying Guan, and Siwei Lyu.
\newblock Glff: Global and local feature fusion for face forgery detection.
\newblock {\em arXiv preprint arXiv:2211.08615}, 2022.

\bibitem{ju2022fusing}
Yan Ju, Shan Jia, Lipeng Ke, Hongfei Xue, Koki Nagano, and Siwei Lyu.
\newblock Fusing global and local features for generalized ai-synthesized image
  detection.
\newblock In {\em 2022 IEEE International Conference on Image Processing
  (ICIP)}, pages 3465--3469. IEEE, 2022.

\bibitem{karras2017progressive}
Tero Karras, Timo Aila, Samuli Laine, and Jaakko Lehtinen.
\newblock Progressive growing of gans for improved quality, stability, and
  variation.
\newblock {\em arXiv preprint arXiv:1710.10196}, 2017.

\bibitem{karras2021alias}
Tero Karras, Miika Aittala, Samuli Laine, Erik H{\"a}rk{\"o}nen, Janne
  Hellsten, Jaakko Lehtinen, and Timo Aila.
\newblock Alias-free generative adversarial networks.
\newblock {\em NeurIPS}, 34, 2021.

\bibitem{karras2019style}
Tero Karras, Samuli Laine, and Timo Aila.
\newblock A style-based generator architecture for generative adversarial
  networks.
\newblock In {\em Proceedings of the IEEE/CVF conference on computer vision and
  pattern recognition}, pages 4401--4410, 2019.

\bibitem{karras2020analyzing}
Tero Karras, Samuli Laine, Miika Aittala, Janne Hellsten, Jaakko Lehtinen, and
  Timo Aila.
\newblock Analyzing and improving the image quality of stylegan.
\newblock In {\em Proceedings of the IEEE/CVF conference on computer vision and
  pattern recognition}, pages 8110--8119, 2020.

\bibitem{kawar2022imagic}
Bahjat Kawar, Shiran Zada, Oran Lang, Omer Tov, Huiwen Chang, Tali Dekel, Inbar
  Mosseri, and Michal Irani.
\newblock Imagic: Text-based real image editing with diffusion models.
\newblock {\em arXiv preprint arXiv:2210.09276}, 2022.

\bibitem{kingma2019introduction}
Diederik~P Kingma, Max Welling, et~al.
\newblock An introduction to variational autoencoders.
\newblock {\em Foundations and Trends{\textregistered} in Machine Learning},
  12(4):307--392, 2019.

\bibitem{kong2022detect}
Chenqi Kong, Baoliang Chen, Haoliang Li, Shiqi Wang, Anderson Rocha, and Sam
  Kwong.
\newblock Detect and locate: Exposing face manipulation by semantic-and
  noise-level telltales.
\newblock {\em IEEE Transactions on Information Forensics and Security},
  17:1741--1756, 2022.

\bibitem{korus2016evaluation}
Pawe{\l} Korus and Jiwu Huang.
\newblock Evaluation of random field models in multi-modal unsupervised
  tampering localization.
\newblock In {\em 2016 IEEE international workshop on information forensics and
  security (WIFS)}, pages 1--6. IEEE, 2016.

\bibitem{2022cat-v2}
Myung-Joon Kwon, Seung-Hun Nam, In-Jae Yu, Heung-Kyu Lee, and Changick Kim.
\newblock Learning jpeg compression artifacts for image manipulation detection
  and localization.
\newblock {\em International Journal of Computer Vision}, 130(8):1875--1895,
  Aug. 2022.

\bibitem{2021cat-v1}
Myung-Joon Kwon, In-Jae Yu, Seung-Hun Nam, and Heung-Kyu Lee.
\newblock Cat-net: Compression artifact tracing network for detection and
  localization of image splicing.
\newblock In {\em Proceedings of the IEEE/CVF Winter Conference on Applications
  of Computer Vision}, pages 375--384, 2021.

\bibitem{lin2014microsoft}
Tsung-Yi Lin, Michael Maire, Serge Belongie, James Hays, Pietro Perona, Deva
  Ramanan, Piotr Doll{\'a}r, and C~Lawrence Zitnick.
\newblock Microsoft coco: Common objects in context.
\newblock In {\em Computer Vision--ECCV 2014: 13th European Conference, Zurich,
  Switzerland, September 6-12, 2014, Proceedings, Part V 13}, pages 740--755.
  Springer, 2014.

\bibitem{lin2023image}
Xun Lin, Shuai Wang, Jiahao Deng, Ying Fu, Xiao Bai, Xinlei Chen, Xiaolei Qu,
  and Wenzhong Tang.
\newblock Image manipulation detection by multiple tampering traces and edge
  artifact enhancement.
\newblock {\em Pattern Recognition}, 133:109026, 2023.

\bibitem{liu2018deep}
Bo Liu and Chi-Man Pun.
\newblock Deep fusion network for splicing forgery localization.
\newblock In {\em proceedings of the European conference on computer vision
  (ECCV) workshops}, pages 0--0, 2018.

\bibitem{liu2020visual}
Fuxiao Liu, Yinghan Wang, Tianlu Wang, and Vicente Ordonez.
\newblock Visual news: Benchmark and challenges in news image captioning.
\newblock {\em arXiv preprint arXiv:2010.03743}, 2020.

\bibitem{liu2022pscc}
Xiaohong Liu, Yaojie Liu, Jun Chen, and Xiaoming Liu.
\newblock Pscc-net: Progressive spatio-channel correlation network for image
  manipulation detection and localization.
\newblock {\em IEEE Transactions on Circuits and Systems for Video Technology},
  32(11):7505--7517, 2022.

\bibitem{mareen2022comprint}
Hannes Mareen, Dante~Vanden Bussche, Fabrizio Guillaro, Davide Cozzolino, Glenn
  Van~Wallendael, Peter Lambert, and Luisa Verdoliva.
\newblock Comprint: Image forgery detection and localization using compression
  fingerprints.
\newblock {\em arXiv preprint arXiv:2210.02227}, 2022.

\bibitem{2020forensicgraphs}
O. {Mayer} and M.~C. {Stamm}.
\newblock Exposing fake images with forensic similarity graphs.
\newblock {\em IEEE Journal of Selected Topics in Signal Processing},
  14(5):1049--1064, 2020.

\bibitem{mokady2022null}
Ron Mokady, Amir Hertz, Kfir Aberman, Yael Pritch, and Daniel Cohen-Or.
\newblock Null-text inversion for editing real images using guided diffusion
  models.
\newblock {\em arXiv preprint arXiv:2211.09794}, 2022.

\bibitem{muller2022diffusion}
Gustav M{\"u}ller-Franzes, Jan~Moritz Niehues, Firas Khader, Soroosh~Tayebi
  Arasteh, Christoph Haarburger, Christiane Kuhl, Tianci Wang, Tianyu Han, Sven
  Nebelung, Jakob~Nikolas Kather, et~al.
\newblock Diffusion probabilistic models beat gans on medical images.
\newblock {\em arXiv preprint arXiv:2212.07501}, 2022.

\bibitem{ng2004data}
Tian-Tsong Ng, Shih-Fu Chang, and Q Sun.
\newblock A data set of authentic and spliced image blocks.
\newblock {\em Columbia University, ADVENT Technical Report}, 4, 2004.

\bibitem{nichol2021glide}
Alex Nichol, Prafulla Dhariwal, Aditya Ramesh, Pranav Shyam, Pamela Mishkin,
  Bob McGrew, Ilya Sutskever, and Mark Chen.
\newblock Glide: Towards photorealistic image generation and editing with
  text-guided diffusion models.
\newblock {\em arXiv preprint arXiv:2112.10741}, 2021.

\bibitem{nichol2021improved}
Alexander~Quinn Nichol and Prafulla Dhariwal.
\newblock Improved denoising diffusion probabilistic models.
\newblock In {\em International Conference on Machine Learning}, pages
  8162--8171. PMLR, 2021.

\bibitem{novozamsky2020imd2020}
Adam Novozamsky, Babak Mahdian, and Stanislav Saic.
\newblock Imd2020: A large-scale annotated dataset tailored for detecting
  manipulated images.
\newblock In {\em Proceedings of the IEEE/CVF Winter Conference on Applications
  of Computer Vision Workshops}, pages 71--80, 2020.

\bibitem{DALLE2ramesh2022hiera}
Aditya Ramesh, Prafulla Dhariwal, Alex Nichol, Casey Chu, and Mark Chen.
\newblock Hierarchical text-conditional image generation with clip latents.
\newblock {\em arXiv preprint arXiv:2204.06125}, 2022.

\bibitem{DALLEramesh2021zero}
Aditya Ramesh, Mikhail Pavlov, Gabriel Goh, Scott Gray, Chelsea Voss, Alec
  Radford, Mark Chen, and Ilya Sutskever.
\newblock Zero-shot text-to-image generation.
\newblock In {\em International Conference on Machine Learning}, pages
  8821--8831. PMLR, 2021.

\bibitem{rombach2022high}
Robin Rombach, Andreas Blattmann, Dominik Lorenz, Patrick Esser, and Bj{\"o}rn
  Ommer.
\newblock High-resolution image synthesis with latent diffusion models.
\newblock In {\em Proceedings of the IEEE/CVF Conference on Computer Vision and
  Pattern Recognition}, pages 10684--10695, 2022.

\bibitem{saharia2022palette}
Chitwan Saharia, William Chan, Huiwen Chang, Chris Lee, Jonathan Ho, Tim
  Salimans, David Fleet, and Mohammad Norouzi.
\newblock Palette: Image-to-image diffusion models.
\newblock In {\em ACM SIGGRAPH 2022 Conference Proceedings}, pages 1--10, 2022.

\bibitem{Imagen_saharia2022phot}
Chitwan Saharia, William Chan, Saurabh Saxena, Lala Li, Jay Whang, Emily
  Denton, Seyed Kamyar~Seyed Ghasemipour, Burcu~Karagol Ayan, S~Sara Mahdavi,
  Rapha~Gontijo Lopes, et~al.
\newblock Photorealistic text-to-image diffusion models with deep language
  understanding.
\newblock {\em arXiv preprint arXiv:2205.11487}, 2022.

\bibitem{shao2023dgm4}
Rui Shao, Tianxing Wu, and Ziwei Liu.
\newblock Detecting and grounding multi-modal media manipulation.
\newblock In {\em IEEE Conference on Computer Vision and Pattern Recognition
  (CVPR)}, 2023.

\bibitem{sinha2021d2c}
Abhishek Sinha, Jiaming Song, Chenlin Meng, and Stefano Ermon.
\newblock D2c: Diffusion-decoding models for few-shot conditional generation.
\newblock {\em Advances in Neural Information Processing Systems},
  34:12533--12548, 2021.

\bibitem{stypulkowski2023diffused}
Micha{\l} Stypu{\l}kowski, Konstantinos Vougioukas, Sen He, Maciej Zi{\k{e}}ba,
  Stavros Petridis, and Maja Pantic.
\newblock Diffused heads: Diffusion models beat gans on talking-face
  generation.
\newblock {\em arXiv preprint arXiv:2301.03396}, 2023.

\bibitem{sun2022lossless}
Chentian Sun, Xiaopeng Fan, and Debin Zhao.
\newblock Lossless recompression of jpeg images using transform domain intra
  prediction.
\newblock {\em IEEE Transactions on Image Processing}, 32:88--99, 2022.

\bibitem{vahdat2021score}
Arash Vahdat, Karsten Kreis, and Jan Kautz.
\newblock Score-based generative modeling in latent space.
\newblock {\em Advances in Neural Information Processing Systems},
  34:11287--11302, 2021.

\bibitem{verdoliva2020media}
Luisa Verdoliva.
\newblock Media forensics and deepfakes: an overview.
\newblock {\em IEEE Journal of Selected Topics in Signal Processing},
  14(5):910--932, 2020.

\bibitem{wang2022objectformer}
Junke Wang, Zuxuan Wu, Jingjing Chen, Xintong Han, Abhinav Shrivastava, Ser-Nam
  Lim, and Yu-Gang Jiang.
\newblock Objectformer for image manipulation detection and localization.
\newblock In {\em Proceedings of the IEEE/CVF Conference on Computer Vision and
  Pattern Recognition}, pages 2364--2373, 2022.

\bibitem{wang2022jpeg}
Menglu Wang, Xueyang Fu, Jiawei Liu, and Zheng-Jun Zha.
\newblock Jpeg compression-aware image forgery localization.
\newblock In {\em Proceedings of the 30th ACM International Conference on
  Multimedia}, pages 5871--5879, 2022.

\bibitem{wang2020cnn}
Sheng-Yu Wang, Oliver Wang, Richard Zhang, Andrew Owens, and Alexei~A Efros.
\newblock Cnn-generated images are surprisingly easy to spot... for now.
\newblock In {\em Proceedings of the IEEE/CVF conference on computer vision and
  pattern recognition}, pages 8695--8704, 2020.

\bibitem{wang2022gan}
Xin Wang, Hui Guo, Shu Hu, Ming-Ching Chang, and Siwei Lyu.
\newblock Gan-generated faces detection: A survey and new perspectives.
\newblock {\em arXiv preprint arXiv:2202.07145}, 2022.

\bibitem{wen2016coverage}
Bihan Wen, Ye Zhu, Ramanathan Subramanian, Tian-Tsong Ng, Xuanjing Shen, and
  Stefan Winkler.
\newblock Coverage—a novel database for copy-move forgery detection.
\newblock In {\em 2016 IEEE international conference on image processing
  (ICIP)}, pages 161--165. IEEE, 2016.

\bibitem{2019ManTraNet}
Yue Wu, Wael AbdAlmageed, and Premkumar Natarajan.
\newblock Mantra-net: Manipulation tracing network for detection and
  localization of image forgeries with anomalous features.
\newblock In {\em Proceedings of the IEEE/CVF Conference on Computer Vision and
  Pattern Recognition}, pages 9543--9552, 2019.

\bibitem{yu2022scaling}
Jiahui Yu, Yuanzhong Xu, Jing~Yu Koh, Thang Luong, Gunjan Baid, Zirui Wang,
  Vijay Vasudevan, Alexander Ku, Yinfei Yang, Burcu~Karagol Ayan, et~al.
\newblock Scaling autoregressive models for content-rich text-to-image
  generation.
\newblock {\em arXiv preprint arXiv:2206.10789}, 2022.

\bibitem{zhou2018learning}
Peng Zhou, Xintong Han, Vlad~I Morariu, and Larry~S Davis.
\newblock Learning rich features for image manipulation detection.
\newblock In {\em Proceedings of the IEEE conference on computer vision and
  pattern recognition}, pages 1053--1061, 2018.

\bibitem{zhou2022detecting}
Xingyi Zhou, Rohit Girdhar, Armand Joulin, Philipp Kr{\"a}henb{\"u}hl, and
  Ishan Misra.
\newblock Detecting twenty-thousand classes using image-level supervision.
\newblock In {\em Computer Vision--ECCV 2022: 17th European Conference, Tel
  Aviv, Israel, October 23--27, 2022, Proceedings, Part IX}, pages 350--368.
  Springer, 2022.

\bibitem{zhou2021lafite}
Yufan Zhou, Ruiyi Zhang, Changyou Chen, Chunyuan Li, Chris Tensmeyer, Tong Yu,
  Jiuxiang Gu, Jinhui Xu, and Tong Sun.
\newblock Lafite: Towards language-free training for text-to-image generation.
\newblock {\em arXiv preprint arXiv:2111.13792}, 2021.

\bibitem{zhuang2022uia}
Wanyi Zhuang, Qi Chu, Zhentao Tan, Qiankun Liu, Haojie Yuan, Changtao Miao,
  Zixiang Luo, and Nenghai Yu.
\newblock Uia-vit: Unsupervised inconsistency-aware method based on vision
  transformer for face forgery detection.
\newblock In {\em Computer Vision--ECCV 2022: 17th European Conference, Tel
  Aviv, Israel, October 23--27, 2022, Proceedings, Part V}, pages 391--407.
  Springer, 2022.

\end{thebibliography}
}

\end{document}